\documentclass[a4paper]{article}

\usepackage{INTERSPEECH2021}

\usepackage{svg}
\usepackage{graphicx,multirow}
\usepackage{hyperref}

\title{Contextual Semi-Supervised Learning: An Approach To Leverage Air-Surveillance and Untranscribed ATC Data in ASR Systems}
\name{Juan Zuluaga-Gomez$^{1,2}$, Iuliia Nigmatulina$^{1}$, Amrutha Prasad$^{1,3}$, Petr Motlicek$^1$, Karel  Vesel\'{y}$^{3}$, Martin Kocour$^{3}$,  Igor Szöke$^{4}$}
\address{
  $^1$Idiap Research Institute, Martigny, Switzerland\\
  $^2$Ecole Polytechnique Federale de Lausanne (EPFL), Switzerland\\
  $^3$Brno University of Technology, Speech@FIT, IT4I CoE, Brno, Czech Republic\\
  $^4$ReplayWell, Brno, Czech Republic
}
\email{\{juan-pablo.zuluaga,iuliia.nigmatulina,aprasad,petr.motlicek\}@idiap.ch,\\
\{iveselyk,ikocour\}@fit.vutbr.cz, szoke@replaywell.com}

\begin{document}

\maketitle

\begin{abstract}
Air traffic management and specifically air-traffic control (ATC) rely mostly on voice communications between Air Traffic Controllers (ATCos) and pilots. In most cases, these voice communications follow a well-defined grammar that could be leveraged in Automatic Speech Recognition (ASR) technologies. The callsign used to address an airplane is an essential part of all ATCo-pilot communications. We propose a two-step approach to add contextual knowledge during semi-supervised training to reduce the ASR system error rates at recognizing the part of the utterance that contains the callsign. Initially, we represent in a WFST the contextual knowledge (i.e. air-surveillance data) of an ATCo-pilot communication. Then, during Semi-Supervised Learning (SSL) the contextual knowledge is added by second-pass decoding (i.e. lattice re-scoring). Results show that `unseen domains' (e.g. data from airports not present in the supervised training data) are further aided by contextual SSL when compared to standalone SSL. For this task, we introduce the Callsign Word Error Rate (CA-WER) as an evaluation metric, which only assesses ASR performance of the spoken callsign in an utterance. We obtained a 32.1\% CA-WER relative improvement applying SSL with an additional 17.5\% CA-WER improvement by adding contextual knowledge during SSL on a challenging ATC-based test set gathered from LiveATC.
\end{abstract}
\noindent\textbf{Index Terms}: automatic speech recognition, contextual semi-supervised learning, air traffic control, air-surveillance data, callsign detection.

\section{Introduction}
\label{sec:introduction}
ATCos regulate and ensure the safety and reliability of air traffic movements by providing spoken guidance to pilots during all flight phases, e.g. approach, landing, taxi, and take-off. This task has been demonstrated to be demanding and stressful~\cite{karlsson1990automatic}. Their most important working tools are their ability to speak articulately, and to master radio, radar, and flight plans. ATC communications follow a well-defined grammar and set of words. However in many cases, there are deviations from the official phraseology in both vocabulary and syntax.
 

Recently, the European Union and Clean Sky Joint Undertaking\footnote{Clean Sky is the largest European research program developing innovative, cutting-edge technology aimed at reducing CO2, gas emissions, and noise levels produced by aircraft. \url{https://www.cleansky.eu}} with the aim of decreasing ATCos workload, increasing air-space safety, and reducing aircraft pollution have been supporting projects such as MALORCA, HAAWAII, and ATCO2, producing detailed results on how to reduce ATCos workload~\cite{helmke2016reducing}, increase their efficiency~\cite{helmke2017increasing}, and how to integrate contextual information in the ASR pipeline~\cite{oualil2015real,shore2012knowledge}. 

Previous research under ATCO2 project targeted cross-accented ASR in ATC, where more than 142 hours from different sources and airports were used for supervised training~\cite{zuluaga2020automatic}. Preliminary results on four test sets suggest that the ASR system can generalize towards speakers with different English accents as long as sufficient amount of manually transcribed training data is available~\cite{zuluaga2020automatic}. In fact, current commercial ASR systems are trained on thousands of annotated speech data whereas in ATC domain not even a considerable fraction of that amount is available for supervised training. Recent research on ASR in ATC has concluded that the lack of annotated speech data and its high production cost are current issues holding the development of fully autonomous ASR systems~\cite{cordero2012automated}.
Some previous research addressed the lack of transcribed ATC speech data using semi-supervised training (e.g. ASR tasks applied to under-resourced languages~\cite{imseng2012comparing,Imseng_ICASSP_2014,dey2019exploiting,khonglah2020incremental}) to decrease Word Error Rates (WER)~\cite{kleinert2018semi,srinivasamurthy2017semi}. In this paper we investigate the effect of integrating contextual knowledge from air-surveillance data into the Semi-Supervised Learning (SSL) pipeline to further boost the performance gains. Detailed information on the proposed approach is given in section \ref{sec:methods}. Similar research adding contextual knowledge into the decoding graph (HCLG.fst) or by re-scoring lattices after the decoding step were described in~\cite{braun2021comparison,aleksic2015bringing,hall2015composition,velikovich2018semantic}. Modifying the Language Model (LM) with prior knowledge is reviewed in~\cite{scheiner2016voice,vasserman2016contextual}.

In Section 2 we present the main ATC task and how SSL and contextual knowledge can be used to leverage the ASR system. Section 3 and Section 4 presents the experimental setup and the two input streams of data, i.e. air-surveillance data and untranscribed ATC voice communications. Section 5 presents the main results and discussion. Finally, Section 6 concludes the paper and proposes a road-map about how to scale up this method for ASR systems trained on data from different airports.

\section{Contextual ASR \& semi-supervised learning}

An ATCo-pilot communication heavily rely on the very particular context they are in. Characteristics such as airplane location, altitude, departure or arrival, and air-space status define the information that could be uttered by the speakers (small deviations are allowed in specific scenarios). For instance, an ASR system can leverage this particular contextual information (mentioned above) as prior knowledge to increase its performance. However, aspects such as speaker's characteristics, location and context, low Signal-to-Noise Ratio (SNR) levels, and air-space status increase the challenge of ASR for the ATC task. 

\subsection{Contextual automatic speech recognition}

Our work relies mostly on adding air-surveillance data as contextual knowledge in the ASR system, also known as `contextual ASR'. Contextual ASR has been an active topic of research in the last decade, where companies such as Google and Microsoft have leveraged contextual data (e.g. user location and contact list) for boosting mobile devices' ASR performance. One of the straightforward ways of adding context into the system is by biasing the LM. In~\cite{aleksic2015bringing}, an on-the-fly re-scoring algorithm allows the insertion of contextual knowledge to the output of the system, with a set of n-grams represented as a \textit{`Biasing'} WFST. Similarly, \cite{hall2015composition} proposes an updated version of the previous biased (\textit{`B'}) WFST. These two previous studies are very related to what we propose in this work. But here, we apply the `biasing' technique in SSL rather than standard ASR training to improve the system's performance. Further research focused on augmenting the n-gram LMs with contextual information (e.g. adjusting the LM probabilities on-the-fly) is reviewed in~\cite{scheiner2016voice} or on injecting classes into a non-class-based LM~\cite{vasserman2016contextual}. In ~\cite{velikovich2018semantic}, the authors explored semantic information inside the decoded word lattice by employing named entity recognition to identify and boost some contextually relevant paths. Finally, research in adding contextual knowledge in end-to-end ASR systems were presented in~\cite{zhao2019shallow}.

\subsection{Contextual ASR in air-traffic control communications}

The International Civil Aviation Organization (ICAO) is the entity that regulates the phraseology and grammar used in ATCo-pilot voice communications. A standard communication starts with a callsign, followed by a command, and a value. One of the main challenges in ATC (thus in ASR) is to correctly identify the inner callsign in the utterance that specifically addresses an individual aircraft. This research focuses on using a list of callsigns as prior knowledge in the ASR system to reduce the search space, thus increasing overall recognition performance. Previous work has attempted to incorporate contextual knowledge in the recognition process. Shore et al.~\cite{shore2012knowledge} targeted word lattice re-scoring with dynamic context (obtained from an independent ATC system that generates a list of possible ``commands") to improve the network recognition performance. Further research on this line of work was presented in \cite{oualil2015real,schmidt2014context,oualil2017context}. We redirect the reader to a general review about spoken instruction understanding in the ATC domain to~\cite{lin2021spoken}. Nevertheless, most of the previously cited works in ASR for ATC employ only data from few airports assuming high-quality speech, i.e. high SNR $\sim$20dB. However, it is hard to determine in advance the quality of each ATCo-pilot communication due to a range of elements such as weather, cockpit or environmental noise.

\subsection{Semi-supervised learning in ASR}

SSL has been proven to be an important asset for ASR in many tasks. The goal of SSL is to leverage large amounts of non-annotated (i.e. data augmented with automatically generated transcripts) data to boost the performance of the ASR trained in an supervised manner. There have been many recent studies leveraging untranscribed data during ASR training; for example, pre-training and self-training methods in end-to-end ASR systems \cite{zhang2020pushing}. Other research has leveraged non-annotated data for ASR in low-resource languages \cite{khonglah2020incremental}. Regarding ATC voice communications, previous researchers have explored different techniques for leveraging untranscribed ATC data with SSL~\cite{kleinert2018semi,srinivasamurthy2017semi}.

\section{Datasets and Methods}
\label{sec:methods}

We propose a method for leveraging contextual data during semi-supervised acoustic model training. In this context, the system is fed with two input streams of data: i)~transcribed and untranscribed ATC voice communications and ii)~corresponding contextual information in the form of air-surveillance data gathered from OpenSky Network (OSN). The air-surveillance contextual data is composed of a list of callsigns for each utterance in the untranscribed dataset. In normal conditions, one of these callsigns is present in the utterance.

\subsection{Supervised ATC databases}

The supervised database is composed of more than 100 hours of mostly clean speech recordings from public domain resources (Atcosim~\cite{ATCOSIM}, UWB\_atcc~\cite{PILSEN_ATC} and LDC\_atcc~\cite{LDC_ATCC}) and from Air Navigation Service Providers (ANSPs) such as in previous projects (Prague and Vienna airports for MALORCA~\cite{kleinert2018semi,srinivasamurthy2017semi} and Toulouse-Blagnac for AIRBUS~\cite{AIRBUS}). The transcripts normalization of these databases was a challenging task due to multiple file formats and annotation ontology. The speakers' accent for each database is country-dependent (e.g. Airbus contains mostly French-accented English recordings). We tested our ASR systems on \textit{Airbus} (1~hr), \textit{Prague} (2.2~hr) and \textit{Vienna} (1.9~hr) test sets, which mostly contain clean speech. Detailed description of these transcribed databases are in~\cite{zuluaga2020automatic, zuluaga2020callsign}.

\subsection{Data from very-high frequency receivers}

There are several ways to obtain untranscribed ATC speech data. For this study we gathered data from two sources that rely on Very-High Frequency (VHF) receivers: i)~open-source channels such as LiveATC\footnote{LiveATC.net is a streaming audio network consisting of local receivers tuned to aircraft communications: https://www.liveatc.net/}, and ii)~recordings from high-quality VHF receivers offered by one project partner (ReplayWell). The recording quality is proportional to how far the VHF receiver is from the speaker (ATCo/pilot) and the hardware quality. First, we manually transcribed 1.9~hours of recordings (mostly noisy speech) from LiveATC to assemble a challenging test set. We tag it as `\textit{liveatc\_mix}' including recordings from EIDW, LSZH, KATL, EHAM, ESGG, and ESOW airports. The SNR levels for \textit{liveatc\_mix} test set ranges from 5-15~dB. Secondly, we gathered 67~hours (49~thousand segments) of ATCo-pilot speech with high-quality setups of VHF receivers in Prague (LKPR) and Brno (LKTB) airports from August 2020 until January 2021. We tag it as `\textit{unsup\_vhf\_67h}' untranscribed train set. We annotated 5 minutes (without silences) of speech collected with VHF receivers from Brno airport (not present in the supervised data), i.e. `\textit{aiport\_lktb\_vhf}' test set. Additionally, we automatically extract timestamp and location information for each utterance in \textit{unsup\_vhf\_67h}. 

\begin{figure}[t]
  \centering
  \includegraphics[width=0.46\textwidth]{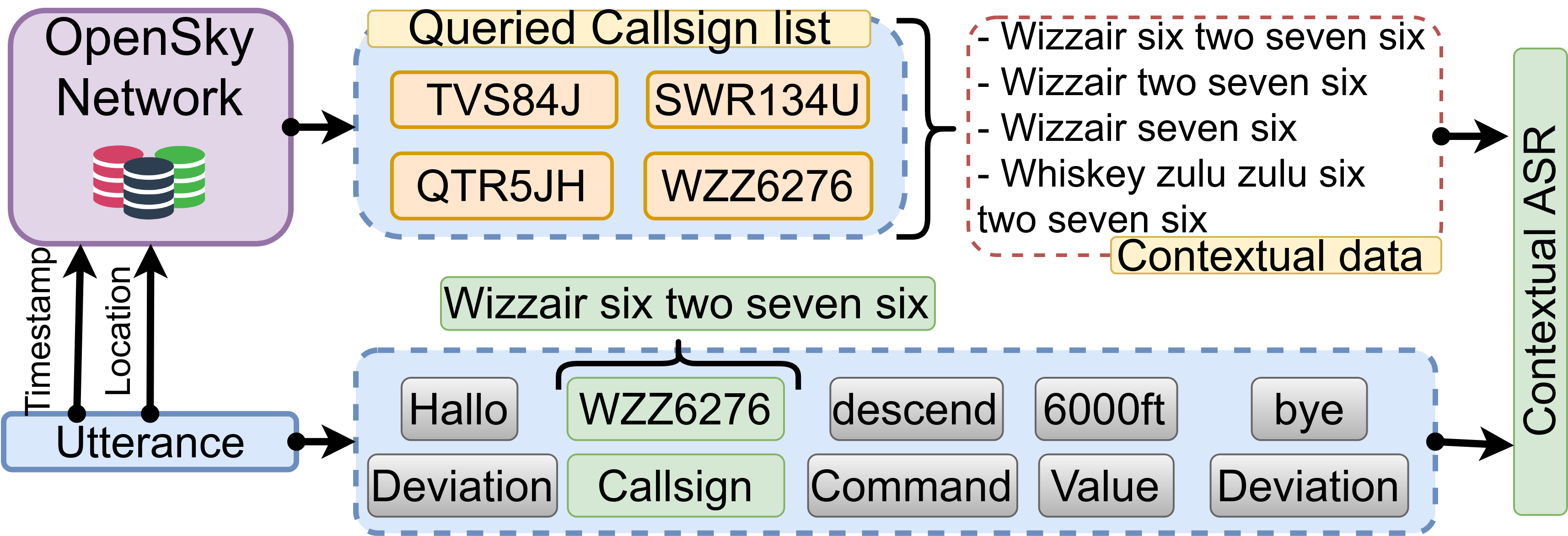}
  \caption{Process of retrieving a list of callsigns (contextual data) from OpenSky Network. The contextual data is the compendium of all possible verbalized versions of each callsign.} 
  \label{fig:osn_query}
\end{figure}

\subsection{Contextual knowledge in semi-supervised learning}

Currently, all the airplanes circulating in Europe must be equipped with Automatic Dependent Surveillance–Broadcast and Mode S modules which transmit almost in real-time their information as meta-data such as altitude, velocity, callsign, and direction. OpenSky Network\footnote{OpenSky Network: provides open access of real-world air traffic control data to the public} captures that information, which can be retrieved by defining a query. We define a query based on the utterances' timestamp and scanned area (\textit{unsup\_vhf\_67h} untranscribed set). OSN retrieves a list of callsigns in ICAO format for each utterance that match the query criteria (potentially one callsign from this list is present in the given utterance). However, our ASR system is trained with transcripts that have the verbalized version of the callsigns instead of ICAO format. We developed an algorithm that verbalizes the ICAO callsigns into different versions. The process is then repeated for each callsign from the callsign list. Figure \ref{fig:osn_query} shows the pipeline to assemble the contextual data from the verbalized callsign list for one utterance. Our previous work can give a more in-depth idea on how the list of callsigns are retrieved and verbalized~\cite{zuluaga2020callsign}. Finally, we repeat this pipeline for each utterance of the unsupervised train set, \textit{unsup\_vhf\_67h}.

\section{Experimental setup}

At this stage, we have the manually transcribed ATC data, five test sets (Airbus, Prague, Vienna, \textit{liveatc\_mix} and aiport\_lktb\_vhf), and 67~hours of untranscribed training data (\textit{unsup\_vhf\_67h}) with their respective contextual information. The experimental setup is divided into: i)~lexicon, language model and seed model training, and ii)~contextual semi-supervised learning. 

\subsection{Baseline ASR system}

The lexicon is composed of a word-list assembled from the transcripts of all available annotated train databases and from additional public resources (e.g. airlines names, airports, countries, ICAO alphabet, way-points, etc). The pronunciation of new words (very common in ATC communications) is obtained with Phonetisaurus G2P \cite{phonetisaurus}. The language model is a tri-gram LM created by interpolating several LMs. An additional LM (only used during the interpolation, to further tune the final LM) is built from external data such as expanded callsigns from 2019\footnote{\url{https://zenodo.org/record/3901482\#.X5cK9k_0m_4}}, expanded runaways (all combinations) and European way-points.

\begin{figure}[t]
  \centering
  \includegraphics[width=0.46\textwidth]{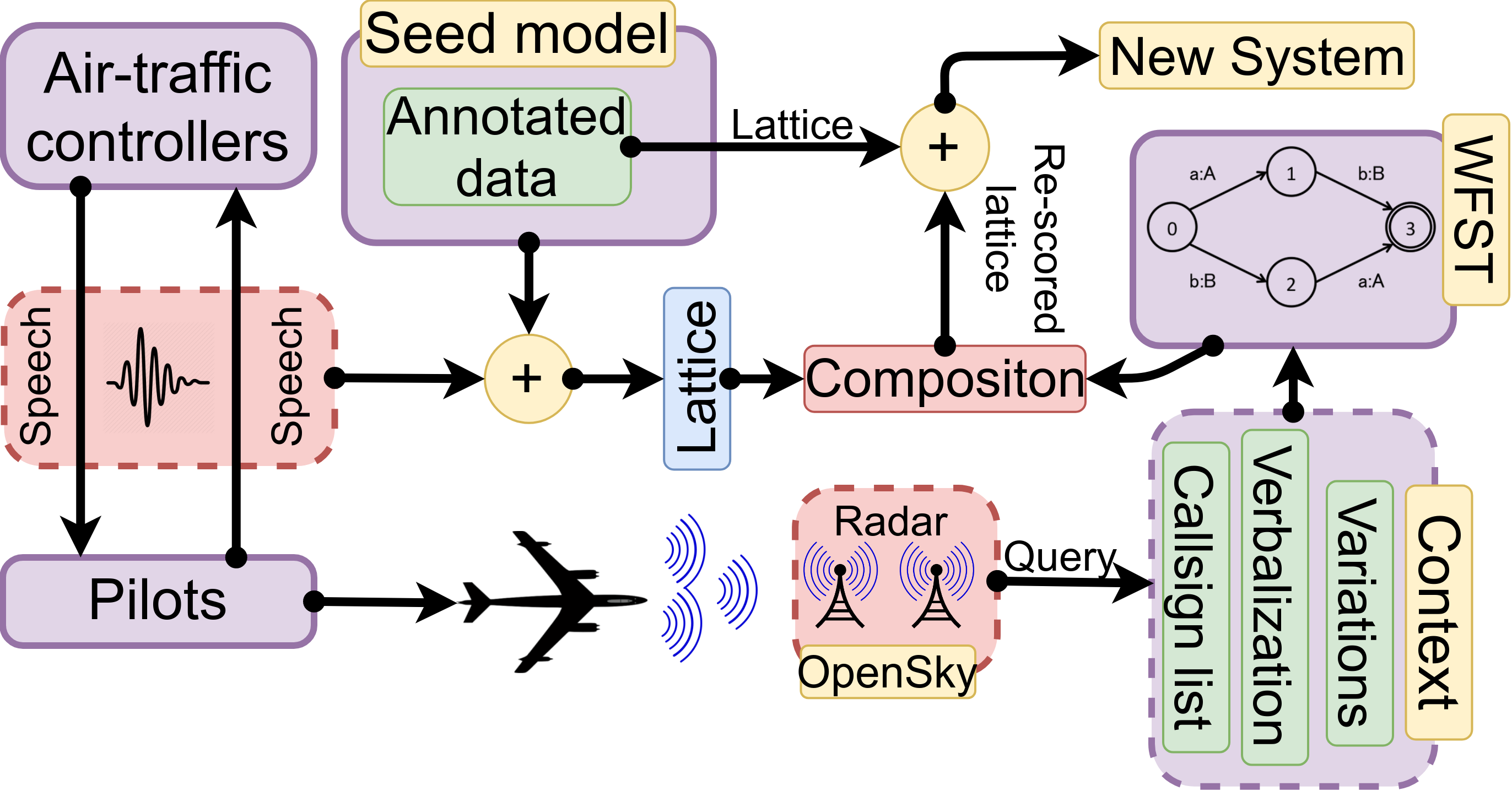}
  \caption{Contextual semi-supervised learning pipeline.} 
  \label{fig:ssl_pipeline}
\end{figure}

All experiments are conducted with Kaldi speech recognition toolkit~\cite{povey2011kaldi}. We report results on state-of-the-art deep neural networks based architectures for hybrid-based ASR. All our models are composed of six convolution layers and 15 factorized time-delay neural network. We use the standard chain Lattice-free MMI (LF-MMI) based Kaldi's recipe~\cite{povey2016purely} for training the seed and SSL-based models. It also requires 100-dimensional i-vector features and 40-dimension MFCC features. Previous experiments~\cite{zuluaga2020callsign} suggested that adding noise to the train data that matches LiveATC audio channel brings considerable improvement in WERs in challenging test sets (e.g. \textit{liveatc\_mix}). We triple the training data by adding noises to the whole database between 5-10dB SNR and then between 10-20dB SNR. The baseline ASR system is trained on all available data for 5 epochs, which we call `seed system'.

\subsection{Contextual semi-supervised learning}

We follow the standard recipe for semi-supervised learning~\cite{khonglah2020incremental}, where a seed system produces word recognition lattices of the untranscribed data set (e.g. \textit{unsup\_vhf\_67h}), which are then mixed with the lattices generated on manually transcribed data to train a new acoustic model. In hybrid ASR, lattices are representations of search results that act as `intermediate format' that contain timing information with more details than plain 1-best string or n-best lists. Lattices generated on manually transcribed and untranscribed data are mixed and a new model is trained with this merged data. There are several ways to add contextual knowledge in the ASR system, e.g. tuning LM towards a defined sequence of n-grams, modifying G.fst when making HCLG graph, or simply re-scoring lattices. This research only explores lattice re-scoring during SSL. 
Initially, we create a Weighted Finite-State Transducer (WFST) graph for each utterance in the untranscribed dataset (i.e \textit{unsup\_vhf\_67h}). The WFST is constructed from n-grams of the verbalized callsign list (air-surveillance data retrieved from OSN). Afterward, the baseline lattices of \textit{unsup\_vhf\_67h} (generated during the first pass decoding) are composed with its particular callsign WFST in a second pass decoding. The re-scored lattices are then used to retrain the acoustic model again as presented in Figure \ref{fig:ssl_pipeline}. In the lattice re-scoring approach, lattices' weights are re-scored to increase the probability of given callsign sequences. The expanded callsigns (represented in WFST) get boosted during the re-scoring process, thus they become more probable to appear in the hypothesized transcripts.

\begin{table}[t!]
  \caption{Word error rates (\%) of several ASR systems for different test sets. The default discount parameter (dp) in ASR systems with lattice re-scoring is 2.0.}
  \label{tab:results}
  \centering
  \begin{tabular}{ p{2.8cm} c c c c c}
  \toprule
    System & {\rotatebox[origin=c]{90}{liveatc\_mix}} & {\rotatebox[origin=c]{90}{aiport\_lktb\_vhf}} & {\rotatebox[origin=c]{90}{Airbus}} & {\rotatebox[origin=c]{90}{Prague}} & {\rotatebox[origin=c]{90}{Vienna}} \\
    \midrule
	Baseline (seed model)   & 49.7  & 26.6  & 11    & 4.4   & 6.8 \\ 
	~~~~~+ SSL                           & 38.3  & 21.3  & 12.1    & 3.8   & 8.2 \\ 
	~~~~~+ lattice re-scoring            & 37.3  & 21.4  & 12.2    & 3.8   & 8.4 \\ 
	\raggedright SSL + lattice re-scoring (dp: 6.0)     & \textbf{36.4}  & \textbf{21.3}  & 11.8    & \textbf{3.6}   & 8.4 \\ 
    \bottomrule
  \end{tabular}
\end{table}

\section{Results and Discussion}

We perform four different experiments to test the proposed approach exploiting the contextual knowledge in SSL (see~Table~\ref{tab:results}). First, we train a baseline acoustic model (i.e. seed model) without semi-supervised learning (first row of Table \ref{tab:results}). Then, we train a new acoustic model from scratch with SSL, the seed model is used to generate the lattices of the untranscribed data set (\textit{unsup\_vhf\_67h}). Next, we re-scored the untranscribed data lattices by composing them with the WFSTs (one for each utterance) previously created. The lattice re-scoring approach relies on a `discount' hyper-parameter, which tells how much weight is given to the `contextual knowledge' encoded at the moment the WFST is created. We report the last result on using a discount parameter of 6.0 instead of 2.0. 

SSL gave much larger improvement for test sets that matched the data used in semi-unsupervised learning (i.e. similar SNR and airport location). For example, we obtained around $\sim$20\% relative WER improvements in \textit{liveatc\_mix} and \textit{aiport\_lktb\_vhf} test sets, and 13.6\% relative WER improvement in Prague test set by doing standalone SSL. Nevertheless, Airbus and Vienna test sets show a WER degradation. We attribute this to data-quality mismatch (i.e. the untranscribed VHF data is noisier than the data with manual transcripts), but also the Airbus and Vienna test sets are from airports not present in the untranscribed set. It is important to mention that WER improvements in challenging test sets such as \textit{liveatc\_mix} and \textit{aiport\_lktb\_vhf} are more significant because the data is nosier and some airports are not present in the annotated train set; which is closer to a real-life scenario. An extra ~$\sim$5\% relative WER improvement is achieved on \textit{liveatc\_mix} and Prague test sets when adding contextual knowledge into the SSL pipeline. The Prague test set yielded improvements in WER in all four proposed ASR systems. We believe this is because data was present in both, the transcribed and untranscribed training sets. 

\begin{figure}[t]
  \centering
  \includegraphics[width=0.46\textwidth]{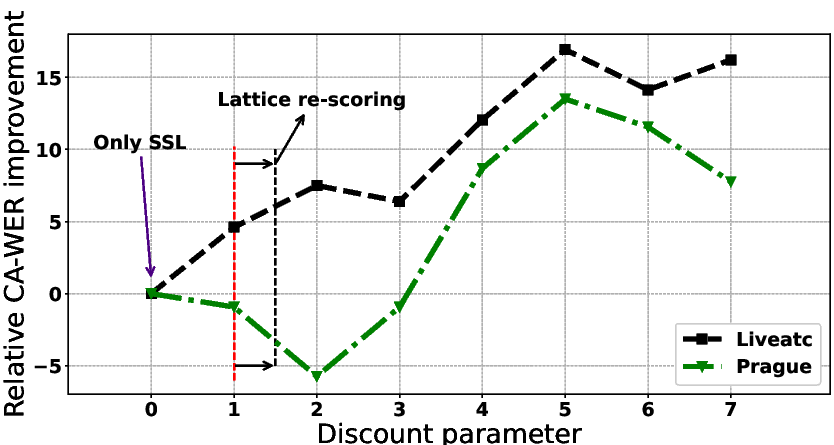}
  \caption{CA-WER performance on liveatc\_mix (noisy) and Prague (clean) test sets for different discount parameters used at the moment of creating the biasing WFST.}
  \label{fig:results}
\end{figure}

The WER metric measures the ASR performance in the whole utterance. Nevertheless, our contextual SSL approach only `boosts' the callsign part in the hypothesized utterance, increasing the chances of recognizing the correct callsign (usually composed of five to seven words, 25\% of the transcript). We therefore propose a new metric: Callsign Word Error Rate `\textit{CA-WER}' which is more aligned to measure the ASR system performance on callsigns only. CA-WER measures only the WER of the callsign between the reference and hypothesized text. We use \textit{texterros}\footnote{https://github.com/RuABraun/texterrors} library to evaluate CA-WER, which needs the verbalized ground truth callsign per utterance. We evaluated CA-WER for \textit{liveatc\_mix}, Prague, and Vienna test sets; 610, 875, and 915 utterances have a callsign, respectively. The CA-WER is evaluated for different discount parameters (hyper-parameter in the WFSTs). Figure \ref{fig:results} shows that lattice re-scoring helps in all cases for \textit{liveatc\_mix} and it helps Prague test set after a discount value of 4.0. Vienna test set is skipped from Figure \ref{fig:results}, because there were no significant variations across different discount parameters. Even though there is a degradation in WER for Vienna test set when adding contextual knowledge, we obtained 7.5\% relative CA-WER improvement when comparing it with the `+ SSL' model (thus showing the robustness of the proposed approach). Discount parameter of 5.0 yielded the best results, reaching a 17.5\% and 14\% CA-WER relative improvement on \textit{liveatc\_mix} (CA-WER:~39.88\% $\rightarrow$ 32.9\%) and Prague (CA-WER:~3.48\%~$\rightarrow$~2.99\%) test sets, respectively (compared to SSL without applying contextual knowledge). 

Finally, the novelty of our approach is that SSL can further leverage contextual knowledge, bypassing the burden of lack of annotated data (which is the case for most of the ATC use-case applications). ATC speech and air-surveillance data can be easily gathered for many airports in Europe, thus the proposed approach could be easily scaled up to different domains/airports. 

\section{Conclusions}

This paper introduced a SSL approach that leverages contextual knowledge. It relies on ATC speech and air-surveillance data. Initially, we create a biasing WFST for each utterance, that encodes n-grams sequences of verbalized callsigns retrieved from OpenSky Network. This prior knowledge in the format of WFST is then added into the SSL recipe to further improve the acoustic models. The WERs did not improve across all cases (test sets) with the proposed approach. However, we obtained significant gains in CA-WER for \textit{liveatc\_mix}, Prague, and Vienna test sets, in comparison to standalone SSL. We believe that CA-WER is a more relevant metric to evaluate the ASR system if we aim to measure its performance regarding `callsign' recognition. Our best ASR system trained with SSL and contextual knowledge yielded a 17\%, 14\% and 7.5\% CA-WER relative improvement in \textit{liveatc\_mix}, Prague, and Vienna test sets compared to standalone semi-supervised learning, respectively. Future research shall explore a better set of discount parameters when building the WFST, for example `rewarding' longer sequences instead of giving the same score for all the boosted sequences. 

\section{Acknowledgements}

The work was supported by the European Union’s Horizon 2020 project No. 864702 - ATCO2 (Automatic collection and processing of voice data from air-traffic communications), which is a part of Clean Sky Joint Undertaking. The work was also partially supported by SESAR Joint Undertaking under HAAWAII project with grant agreement No. 884287.

\bibliographystyle{IEEEtran}

\bibliography{main}

\begin{thebibliography}{10}
\providecommand{\url}[1]{#1}
\csname url@samestyle\endcsname
\providecommand{\newblock}{\relax}
\providecommand{\bibinfo}[2]{#2}
\providecommand{\BIBentrySTDinterwordspacing}{\spaceskip=0pt\relax}
\providecommand{\BIBentryALTinterwordstretchfactor}{4}
\providecommand{\BIBentryALTinterwordspacing}{\spaceskip=\fontdimen2\font plus
\BIBentryALTinterwordstretchfactor\fontdimen3\font minus
  \fontdimen4\font\relax}
\providecommand{\BIBforeignlanguage}[2]{{%
\expandafter\ifx\csname l@#1\endcsname\relax
\typeout{** WARNING: IEEEtran.bst: No hyphenation pattern has been}%
\typeout{** loaded for the language `#1'. Using the pattern for}%
\typeout{** the default language instead.}%
\else
\language=\csname l@#1\endcsname
\fi
#2}}
\providecommand{\BIBdecl}{\relax}
\BIBdecl

\bibitem{karlsson1990automatic}
J.~Karlsson, ``Automatic speech recognition in air traffic control: A human
  factors perspective,'' \emph{Military and Government Speech Technology},
  vol.~1, pp. 13--15, 1989.

\bibitem{helmke2016reducing}
H.~Helmke, O.~Ohneiser, T.~M{\"u}hlhausen, and M.~Wies, ``Reducing controller
  workload with automatic speech recognition,'' in \emph{2016 IEEE/AIAA 35th
  Digital Avionics Systems Conference (DASC)}.\hskip 1em plus 0.5em minus
  0.4em\relax IEEE, 2016, pp. 1--10.

\bibitem{helmke2017increasing}
H.~Helmke, O.~Ohneiser, J.~Buxbaum, and C.~Kern, ``Increasing atm efficiency
  with assistant based speech recognition,'' in \emph{Proc. of the 13th
  USA/Europe Air Traffic Management Research and Development Seminar, Seattle,
  USA}, 2017.

\bibitem{oualil2015real}
Y.~Oualil, M.~Schulder, H.~Helmke, A.~Schmidt, and D.~Klakow, ``Real-time
  integration of dynamic context information for improving automatic speech
  recognition,'' in \emph{Sixteenth Annual Conference of the International
  Speech Communication Association}, 2015.

\bibitem{shore2012knowledge}
T.~Shore, F.~Faubel, H.~Helmke, and D.~Klakow, ``Knowledge-based word lattice
  rescoring in a dynamic context,'' in \emph{Thirteenth Annual Conference of
  the International Speech Communication Association}, 2012.

\bibitem{zuluaga2020automatic}
\BIBentryALTinterwordspacing
J.~Zuluaga{-}Gomez, P.~Motl{\'{\i}}cek, Q.~Zhan, K.~Vesel{\'{y}}, and R.~Braun,
  ``Automatic speech recognition benchmark for air-traffic communications,'' in
  \emph{Interspeech 2020, 21st Annual Conference of the International Speech
  Communication Association, Virtual Event, Shanghai, China, 25-29 October
  2020}.\hskip 1em plus 0.5em minus 0.4em\relax {ISCA}, 2020, pp. 2297--2301.
  [Online]. Available: \url{https://doi.org/10.21437/Interspeech.2020-2173}
\BIBentrySTDinterwordspacing

\bibitem{cordero2012automated}
J.~M. Cordero, M.~Dorado, and J.~M. de~Pablo, ``Automated speech recognition in
  atc environment,'' in \emph{Proceedings of the 2nd International Conference
  on Application and Theory of Automation in Command and Control Systems},
  2012, pp. 46--53.

\bibitem{imseng2012comparing}
D.~Imseng, J.~Dines, P.~Motlicek, P.~N. Garner, and H.~Bourlard, ``Comparing
  different acoustic modeling techniques for multilingual boosting,'' in
  \emph{Thirteenth Annual Conference of the International Speech Communication
  Association}, 2012.

\bibitem{Imseng_ICASSP_2014}
D.~Imseng, B.~Potard, P.~Motlicek, A.~Nanchen, and H.~Bourlard, ``Exploiting
  un-transcribed foreign data for speech recognition in well-resourced
  languages,'' in \emph{Proceedings IEEE International Conference on Acoustics,
  Speech and Signal Processing}.\hskip 1em plus 0.5em minus 0.4em\relax IEEE,
  2014, pp. 2322 -- 2326.

\bibitem{dey2019exploiting}
S.~Dey, P.~Motlicek, T.~Bui, and F.~Dernoncourt, ``Exploiting semi-supervised
  training through a dropout regularization in end-to-end speech recognition,''
  \emph{Proc. Interspeech 2019}, pp. 734--738, 2019.

\bibitem{khonglah2020incremental}
B.~Khonglah, S.~Madikeri, S.~Dey, H.~Bourlard, P.~Motlicek, and J.~Billa,
  ``Incremental semi-supervised learning for multi-genre speech recognition,''
  in \emph{ICASSP 2020-2020 IEEE International Conference on Acoustics, Speech
  and Signal Processing (ICASSP)}.\hskip 1em plus 0.5em minus 0.4em\relax IEEE,
  2020, pp. 7419--7423.

\bibitem{kleinert2018semi}
M.~Kleinert, H.~Helmke, G.~Siol, H.~Ehr, A.~Cerna, C.~Kern, D.~Klakow,
  P.~Motlicek, Y.~Oualil, M.~Singh \emph{et~al.}, ``Semi-supervised adaptation
  of assistant based speech recognition models for different approach areas,''
  in \emph{37th Digital Avionics Systems Conference (DASC)}.\hskip 1em plus
  0.5em minus 0.4em\relax IEEE, 2018, pp. 1--10.

\bibitem{srinivasamurthy2017semi}
A.~Srinivasamurthy, P.~Motlicek, I.~Himawan, G.~Szaszak, Y.~Oualil, and
  H.~Helmke, ``Semi-supervised learning with semantic knowledge extraction for
  improved speech recognition in air traffic control,'' in \emph{Proc. of the
  18th Annual Conference of the International Speech Communication
  Association}, 2017.

\bibitem{braun2021comparison}
R.~A. Braun, S.~Madikeri, and P.~Motlicek, ``A comparison of methods for
  oov-word recognition on a new public dataset,'' in \emph{ICASSP 2021-2021
  IEEE International Conference on Acoustics, Speech and Signal Processing
  (ICASSP)}.\hskip 1em plus 0.5em minus 0.4em\relax IEEE, 2021, pp. 5979--5983.

\bibitem{aleksic2015bringing}
P.~Aleksic, M.~Ghodsi, A.~Michaely, C.~Allauzen, K.~Hall, B.~Roark, D.~Rybach,
  and P.~Moreno, ``Bringing contextual information to google speech
  recognition,'' in \emph{Sixteenth Annual Conference of the International
  Speech Communication Association}, 2015.

\bibitem{hall2015composition}
K.~Hall, E.~Cho, C.~Allauzen, F.~Beaufays, N.~Coccaro, K.~Nakajima, M.~Riley,
  B.~Roark, D.~Rybach, and L.~Zhang, ``Composition-based on-the-fly rescoring
  for salient n-gram biasing,'' in \emph{Sixteenth Annual Conference of the
  International Speech Communication Association}, 2015.

\bibitem{velikovich2018semantic}
L.~Velikovich, I.~Williams, J.~Scheiner, P.~S. Aleksic, P.~J. Moreno, and
  M.~Riley, ``Semantic lattice processing in contextual automatic speech
  recognition for google assistant.'' in \emph{Interspeech}, 2018, pp.
  2222--2226.

\bibitem{scheiner2016voice}
J.~Scheiner, I.~Williams, and P.~Aleksic, ``Voice search language model
  adaptation using contextual information,'' in \emph{2016 IEEE Spoken Language
  Technology Workshop (SLT)}.\hskip 1em plus 0.5em minus 0.4em\relax IEEE,
  2016, pp. 253--257.

\bibitem{vasserman2016contextual}
L.~Vasserman, B.~Haynor, and P.~Aleksic, ``Contextual language model adaptation
  using dynamic classes,'' in \emph{2016 IEEE Spoken Language Technology
  Workshop (SLT)}.\hskip 1em plus 0.5em minus 0.4em\relax IEEE, 2016, pp.
  441--446.

\bibitem{zhao2019shallow}
D.~Zhao, T.~N. Sainath, D.~Rybach, P.~Rondon, D.~Bhatia, B.~Li, and R.~Pang,
  ``Shallow-fusion end-to-end contextual biasing,'' \emph{Proc. Interspeech
  2019}, pp. 1418--1422, 2019.

\bibitem{schmidt2014context}
A.~Schmidt, Y.~Oualil, O.~Ohneiser, M.~Kleinert, M.~Schulder, A.~Khan,
  H.~Helmke, and D.~Klakow, ``Context-based recognition network adaptation for
  improving on-line asr in air traffic control,'' in \emph{2014 IEEE Spoken
  Language Technology Workshop (SLT)}.\hskip 1em plus 0.5em minus 0.4em\relax
  IEEE, 2014, pp. 13--18.

\bibitem{oualil2017context}
Y.~Oualil, D.~Klakow, G.~Szasz{\'a}k, A.~Srinivasamurthy, H.~Helmke, and
  P.~Motlicek, ``A context-aware speech recognition and understanding system
  for air traffic control domain,'' in \emph{2017 IEEE Automatic Speech
  Recognition and Understanding Workshop (ASRU)}.\hskip 1em plus 0.5em minus
  0.4em\relax IEEE, 2017, pp. 404--408.

\bibitem{lin2021spoken}
Y.~Lin, ``Spoken instruction understanding in air traffic control: Challenge,
  technique, and application,'' \emph{Aerospace}, vol.~8, no.~3, p.~65, 2021.

\bibitem{zhang2020pushing}
Y.~Zhang, J.~Qin, D.~S. Park, W.~Han, C.-C. Chiu, R.~Pang, Q.~V. Le, and Y.~Wu,
  ``Pushing the limits of semi-supervised learning for automatic speech
  recognition,'' \emph{arXiv preprint arXiv:2010.10504}, 2020.

\bibitem{ATCOSIM}
K.~Hofbauer, S.~Petrik, and H.~Hering, ``The atcosim corpus of non-prompted
  clean air traffic control speech.'' in \emph{LREC}, 2008.

\bibitem{PILSEN_ATC}
L.~{\v{S}}m{\'\i}dl, J.~{\v{S}}vec, D.~Tihelka, J.~Matou{\v{s}}ek, J.~Romportl,
  and P.~Ircing, ``Air traffic control communication ({ATCC}) speech corpora
  and their use for {ASR} and {TTS} development,'' \emph{Language Resources and
  Evaluation}, vol.~53, no.~3, pp. 449--464, 2019.

\bibitem{LDC_ATCC}
\BIBentryALTinterwordspacing
J.~Godfrey, ``{The Air Traffic Control Corpus (ATC0) - LDC94S14A},'' 1994.
  [Online]. Available: \url{https://catalog.ldc.upenn.edu/LDC94S14A}
\BIBentrySTDinterwordspacing

\bibitem{AIRBUS}
E.~Delpech, M.~Laignelet, C.~Pimm, C.~Raynal, M.~Trzos, A.~Arnold, and
  D.~Pronto, ``{A Real-life, French-accented Corpus of Air Traffic Control
  Communications},'' in \emph{Proceedings of the Eleventh International
  Conference on Language Resources and Evaluation (LREC 2018)}, 2018.

\bibitem{zuluaga2020callsign}
J.~Zuluaga-Gomez, K.~Vesel{\`y}, A.~Blatt, P.~Motlicek, D.~Klakow, A.~Tart,
  I.~Sz{\"o}ke, A.~Prasad, S.~Sarfjoo, P.~Kol{\v{c}}{\'a}rek \emph{et~al.},
  ``Automatic call sign detection: Matching air surveillance data with air
  traffic spoken communications,'' in \emph{Multidisciplinary Digital
  Publishing Institute Proceedings}, vol.~59, no.~1, 2020, p.~14.

\bibitem{phonetisaurus}
J.~R. Novak, N.~Minematsu, and K.~Hirose, ``Phonetisaurus: Exploring
  grapheme-to-phoneme conversion with joint n-gram models in the {WFST}
  framework,'' \emph{Nat. Lang. Eng.}, vol.~22, no.~6, pp. 907--938, 2016.

\bibitem{povey2011kaldi}
D.~Povey, A.~Ghoshal, G.~Boulianne, L.~Burget, O.~Glembek, N.~Goel,
  M.~Hannemann, P.~Motlicek, Y.~Qian, P.~Schwarz \emph{et~al.}, ``The kaldi
  speech recognition toolkit,'' in \emph{IEEE workshop on automatic speech
  recognition and understanding}, no. CONF.\hskip 1em plus 0.5em minus
  0.4em\relax IEEE Signal Processing Society, 2011.

\bibitem{povey2016purely}
D.~Povey, V.~Peddinti, D.~Galvez, P.~Ghahremani, V.~Manohar, X.~Na, Y.~Wang,
  and S.~Khudanpur, ``Purely sequence-trained neural networks for asr based on
  lattice-free mmi.'' in \emph{Interspeech}, 2016, pp. 2751--2755.

\end{thebibliography}

\end{document}